\documentclass[sigconf]{acmart}

\AtBeginDocument{%
  }

\setcopyright{acmlicensed}
\copyrightyear{2018}
\acmYear{2018}
\acmDOI{XXXXXXX.XXXXXXX}
\acmConference[OARS '26]{RecSys '26 OARS workshop}{2026}{Minneapolis, MN, USA}

\acmISBN{978-1-4503-XXXX-X/2018/06}

\begin{document}

\title{Billion-Scale Thumbnail Optimization for Uncurated Short-Form Videos via Multi-Armed Bandits}

\author{Ying Han}
\authornote{Both authors contributed equally to this paper.}
\email{ylhan@google.com}
\affiliation{%
  \institution{Google}
  \city{Mountain View}
  \state{California}
  \country{USA}
}

\author{Ling Liu}
\authornotemark[1]
\email{liuling@google.com}
\affiliation{%
  \institution{Google}
  \city{Mountain View}
  \state{California}
  \country{USA}
}

\author{Fabio Soldo}
\email{fsoldo@google.com}
\affiliation{%
  \institution{Google}
  \city{Mountain View}
  \state{California}
  \country{USA}
}

\author{Vu Nguyen}
\email{vuanguyen@google.com}
\affiliation{%
  \institution{Google}
  \city{Mountain View}
  \state{California}
  \country{USA}
}

\author{Danio Wang}
\email{daniojianxiong@google.com}
\affiliation{%
  \institution{Google}
  \city{Mountain View}
  \state{California}
  \country{USA}
}

\author{Liz Kidd}
\email{lizkidd@google.com}
\affiliation{%
  \institution{Google}
  \city{Mountain View}
  \state{California}
  \country{USA}
}

\author{Yongle Cao}
\email{yonglec@google.com}
\affiliation{%
  \institution{Google}
  \city{Mountain View}
  \state{California}
  \country{USA}
}

\author{Theodore Rose}
\email{tedrose@google.com}
\affiliation{%
  \institution{Google}
  \city{Mountain View}
  \state{California}
  \country{USA}
}

\author{Su-Lin Wu}
\email{sulin@google.com}
\affiliation{%
  \institution{Google}
  \city{Mountain View}
  \state{California}
  \country{USA}
}

\author{Romer Rosales}
\email{romerrosales@google.com}
\affiliation{%
  \institution{Google}
  \city{Mountain View}
  \state{California}
  \country{USA}
}

\renewcommand{\shortauthors}{Y. Han, L. Liu et al.}

\begin{abstract}
This paper introduces a real-time thumbnail optimization system deployed at a global $O(B)$ scale on a major short-form video platform. Unlike traditional long-form content, where custom thumbnails are heavily curated by creators, a considerable fraction of short-form videos are published without human-selected artwork. To address this uncurated corpus, we present a fully automated, end-to-end framework that replaces static default frames with dynamic, data-driven selections across billions of videos. To the best of our knowledge, this is the first published work demonstrating an online Multi-Armed Bandit framework successfully deployed at an $O(B)$ scale for uncurated short-form video discovery. Our solution pairs a multi-stage candidate generation pipeline with a low-latency serving infrastructure. By initializing the exploration framework with image-specific priors derived from a deep visual quality model, the system minimizes exploration costs and dynamically serves optimal thumbnails at serving time. Global deployment demonstrates statistically significant improvements in core user discovery and engagement metrics.

\end{abstract}



\maketitle

\section{Introduction}
Thumbnails serve as the primary visual gateway for content discovery on video-sharing platforms, playing a critical role in guiding user consumption choices \cite{Amat2018}. On browse surfaces such as the homepage, where a multitude of content options compete for viewer attention \cite{Covington2016}, an effective thumbnail significantly influences decision-making and downstream engagement. While short-form video has emerged as a dominant and rapidly growing segment of media consumption \cite{NRG2024Empowering, Deloitte2025DigitalMediaTrends}, it presents highly distinct visual discovery challenges compared to traditional long-form media.

Specifically, short-form video platforms ingest corpora of massive scale where a considerable fraction of uploads lack custom, creator-provided artwork.Historically, the platform addressed this asset gap by serving a single static thumbnail extracted from frames in the middle section of the video. Relying on a single predetermined frame frequently fails to capture the most compelling visual hook of short-form content, leaving substantial headroom for optimization. Preliminary research and online experiments confirm that optimizing thumbnail selection drives highly measurable gains in user value metrics \cite{Amat2018}.

To overcome the strict limitations of static defaults, we transitioned to a dynamic, online learning framework by formulating thumbnail optimization as a Multi-Armed Bandit (MAB) problem \cite{Lattimore2020}. In this setup, high-quality candidate frames extracted from a video represent independent arms. Pure online exploration at this magnitude, however, incurs prohibitive costs by serving sub-optimal frames during the learning phase. To solve this, our architecture extracts features via a specialized deep computer vision model to establish informative visual priors. By guiding probabilistic arm selection with these image-specific priors, our system intelligently explores candidate frames while rapidly exploiting real-time user feedback to converge on the most appealing representation for each video.

To the best of our knowledge, this represents the first published work demonstrating successful industry deployment of real-time probabilistic bandit optimization—guided by deep visual priors—to solve the uncurated thumbnail problem at an $O(B)$ scale. This online adaptation moves beyond static assumptions, aligning visual presentation directly with actual viewer preferences under strict millisecond serving constraints.

The main contributions of this work are threefold:
\begin{enumerate}
    \item A fully automated candidate selection framework tailored for uncurated short-form videos, leveraging a specialized deep computer vision model to extract and score high-quality thumbnail options natively from source frames;
    \item \begin{sloppypar} A real-time, low-latency bandit infrastructure enabling dynamic online selection, uniquely utilizing offline model scores as informative priors to efficiently optimize the exploration phase during cold starts; \end{sloppypar}
    \item Successful global deployment at an $O(B)$ scale, providing empirical evidence of statistically significant, platform-wide improvements in user discovery visits and core engagement metrics.
\end{enumerate}

\section{Related Work}

\subsection{Video Thumbnail Selection}
The problem of video thumbnail selection has been extensively studied, transitioning from heuristic-based keyframe extraction to sophisticated deep learning frameworks. Early methodologies predominantly focused on offline static extraction, relying on basic visual heuristics such as motion blur, contrast, and facial detection to identify appealing frames. For instance, \cite{Song2016} proposed an automated system that leverages objective visual quality metrics (e.g., texture, sharpness, and luminance) and clustering techniques to extract aesthetically pleasing and representative frames.

With the advent of deep learning, multi-modal semantic approaches became the new standard for assessing visual quality and relevance natively from video signals. \cite{Liu2015} introduced a multi-task deep visual-semantic embedding model that selects thumbnails by mapping visual content and side semantic information (e.g., video titles or search queries) into a shared latent space. To eliminate the need for extensive manual annotations, more recent works have framed offline thumbnail selection as a reinforcement learning (RL) problem. \cite{Apostolidis2021} utilized combinations of adversarial training and RL to sequentially select candidate frames that maximize tailored reward functions representing aesthetics and visual representativeness. While these offline methodologies successfully quantify visual and semantic appeal to parse out high-quality images, they remain static post-deployment; they extract a single best frame prior to serving and cannot adapt dynamically to actual, real-time viewer engagement.

\subsection{Dynamic Optimization via Multi-Armed Bandits}

To address the limitations of static offline selection, the industry has increasingly adopted dynamic thumbnail optimization frameworks. A notable pioneer in this space is Netflix \cite{chandrashekar2017artwork, Amat2018}, which demonstrated the evolution of thumbnail optimization from global, unpersonalized multi-armed bandits to contextual bandits that dynamically select artwork tailored to individual user context. Similarly, gaming platforms such as Roblox \cite{RobloxThumbnail2024} have recently introduced dynamic thumbnail testing capabilities, automatically serving multiple creator-provided visual assets to optimize qualified play through rates based on aggregated viewer engagement.

However, these existing systems typically operate over bounded corpora utilizing pre-approved, creator-curated image assets. Extending online learning frameworks to a massive uncurated short-form video corpora introduces unprecedented challenges: first, generating a high-quality action space from raw video frames at scale; second, operating distributed serving at an $O(B)$ scale with strict latency limits; and third, overcoming cold-start exploration penalties.

Our work bridges the gap between these two distinct domains through a holistic, end-to-end framework. First, we employ a highly scalable, multi-stage offline candidate generation pipeline---building on visual quality assessment---to distill raw video frames into a constrained, highly optimized set of candidate arms during ingestion. Second, we introduce an effective online Thompson Sampling architecture capable of continuously operating over billions of these uncurated items in real time. Finally, to further mitigate the cold-start exploration penalty inherent in such a massive deployment, we reuse the offline computer vision model's quality scores to inject informative visual priors into the bandit initialization. This combination of automated candidate curation, robust distributed serving, and intelligent prior initialization enables the effective optimization of global thumbnail discovery at an unprecedented $O(B)$ scale.

\section{Methodology}

\subsection{Method Overview}
Selecting the optimal thumbnail for an uncurated short-form video can be naturally formulated as a Multi-Armed Bandit (MAB) problem. For a given video upload, each candidate frame extracted by the upstream processing pipeline represents an independent arm. The objective of the online algorithm is to identify and dynamically serve the single candidate frame that maximizes the expected probability of positive user engagement across the platform's global audience.

To solve this online exploration-exploitation problem at scale, we evaluated Upper Confidence Bound (UCB) algorithms against Thompson Sampling (TS). While UCB provides robust asymptotic guarantees, its deterministic nature introduces severe operational bottlenecks in a highly concurrent, distributed serving environment. Under massive request volumes, attribution feedback is inherently delayed by streaming log ingestion pipelines. The delayed feedback severely limits the performance of standard UCB policies \cite{joulani2013online}, as the deterministic argmax hands every concurrent impression the same frame until new feedback is logged, collapsing exploration during traffic spikes. 

We selected Thompson Sampling (TS) \cite{Chapelle2011, russo2018tutorial} because its probabilistic serving paradigm is well-suited to address the challenges of distributed latency and cold-start exploration. By sampling directly from the posterior reward distribution of each candidate frame, TS naturally diversifies exploration across concurrent requests even when feedback attribution is delayed. Crucially, the Bayesian foundation of TS enables our core system innovation: the seamless injection of informative deep visual priors. Rather than initializing exploration from a state of uniform uncertainty, our architecture maps the scalar quality scores produced by an offline deep computer vision model directly into the bandit's initial prior distributions. This synthesis allows the system to explore and exploit discriminately among different thumbnail candidates from the very first impression, drastically reducing the user-experience penalty typically associated with cold-start exploration on uncurated corpora.

\subsection{Problem Formulation}
We model user interaction with a candidate thumbnail as a Bernoulli trial, where the reward outcome is binary: success ($Y=1$, representing a qualified discovery visit or sustained playback) or failure ($Y=0$, representing an immediate skip or ignored impression). 

\subsubsection{Notation}
\begin{itemize}
    \item $K$ denotes the total number of candidate thumbnails extracted for a given video $V$.
    \item $\theta_k \in [0, 1]$ is the latent success probability (reward parameter) for candidate thumbnail $T_k$, modeled as a random variable following a Beta distribution.
    \item $Y_k \in \{0, 1\}$ is the observed binary reward action.
    \item $\alpha_k, \beta_k$ are the shape hyperparameters defining the Beta distribution of $\theta_k$.
\end{itemize}

Because the reward distribution is Bernoulli, the Beta distribution serves as its conjugate prior. This ensures that the posterior distribution remains closed-form within the Beta family, enabling highly efficient, lightweight parameter updates under strict millisecond serving constraints. The end-to-end operational lifecycle for a video $V$ proceeds as follows:

\subsubsection{Prior Initialization (Solving the Cold-Start)} 
    For each candidate thumbnail $T_k$, we initialize the \textit{a priori} probability density function $p(\theta_k)$. A standard uninformative setup sets $\alpha_{k,0} = \beta_{k,0} = 1$, yielding a uniform distribution $\text{Beta}(1, 1)$. To eliminate pure-exploration waste, we parameterize the initial pseudo-counts using the candidate's offline visual quality score $q_k \in [0, 1]$ derived from our deep computer vision model. By employing a discretized prior initialization strategy by partitioning the the continuous visual quality score $q_k \in [0,1]$ into a set of disjoint intervals, where each bucket B is mapped to a specific pair of priors $(\alpha_B, \beta_B)$. The priors for a given bucket are derived from the historical mean reward rate $\mu_B $ and the effective sample size $N_{\text{eff}, B}$ that represents the target confidence level for the priors:

    \begin{equation}
    \alpha_{B}= \mu_B \cdot N_{\text{eff}, B}
    \end{equation}
    
    \begin{equation}
    \beta_{B} = (1 - \mu_B) \cdot N_{\text{eff}, B}
    \end{equation}
    This directly skews the initial sampling distribution toward frames with high estimated visual appeal while preserving exploration variance.
    
\subsubsection{Continuous Serving Loop}
    \begin{itemize}
        \item \textbf{Step 1: Arm Selection.} At serving time, sample a predicted success rate $\tilde{\theta}_k$ from the current distribution for each candidate:
        \begin{equation}
            \tilde{\theta}_k \sim \text{Beta}(\alpha_k, \beta_k)
        \end{equation}
        Serve the candidate thumbnail $T^*$ that maximizes the sampled value: $T^* = \arg\max_k \tilde{\theta}_k$.
        
        \item \textbf{Step 2: Telemetry Logging.} Observe the user interaction. Let $n_k$ be the cumulative impressions and $y_k$ be the cumulative reward actions observed for arm $k$. Upon processing a stream batch, increment $n_k \mathrel{+}= 1$; if the impression yielded a reward, increment $y_k \mathrel{+}= 1$.
        
        \item \textbf{Step 3: Posterior Update.} Update the latent parameter distribution using Bayes' Rule:
        \begin{equation}
            p(\theta_k \mid y_k, n_k) = \frac{p(y_k \mid \theta_k, n_k) \cdot p(\theta_k)}{p(y_k)}
        \end{equation}
    \end{itemize}

\subsubsection{Posterior Derivation}
\begin{sloppypar} 
Dropping the arm subscript $k$ for brevity, let the \textit{a priori} distribution be modeled as $p(\theta) \sim \text{Beta}(\alpha, \beta)$. Given $y$ observed successes over $n$ independent Bernoulli trials, the likelihood function follows a Binomial distribution: $p(y \mid \theta, n) = \binom{n}{y} \theta^y (1-\theta)^{n-y}$. 
\end{sloppypar}

The analytic derivation of the \textit{a posterior} distribution updates as follows:

\begin{align}
 p(\theta \mid y, n) &= \frac{p(y \mid \theta, n) \cdot p(\theta)}{p(y)} \nonumber \\
 &\propto p(y \mid \theta, n) \cdot p(\theta) \nonumber \\
 &= \left[ \binom{n}{y} \theta^y (1-\theta)^{n-y} \right] \cdot \left[ \frac{1}{\mathrm{B}(\alpha, \beta)} \theta^{\alpha-1}(1-\theta)^{\beta-1} \right] \nonumber \\
 &\propto \theta^{\alpha+y-1}(1-\theta)^{\beta+n-y-1} \nonumber \\
 &\sim \text{Beta}(\alpha+y, \beta+n-y)
\end{align}

where $\mathrm{B}(\alpha, \beta)$ is the standard Beta function acting as a normalization constant. Because the updated posterior parameters resolve simply to additive updates ($\alpha_{\text{new}} = \alpha + y$ and $\beta_{\text{new}} = \beta + n - y$), the platform scales this inference continuously across $O(B)$ concurrent items with minimal computational overhead.

\section{System Components}

\subsection{Thumbnail Candidate Selection}
The candidate selection phase serves as a critical upstream component designed to extract a visually diverse, highly optimized set of candidate frames to feed the downstream multi-armed bandit (MAB) infrastructure. To explicitly preserve creator intent, this pipeline selectively processes only uploads lacking custom, human-curated artwork. The end-to-end multi-stage generation workflow is illustrated in Figure 1.

\begin{figure*}[h]
  \centering
  \includegraphics[width=0.9\linewidth]{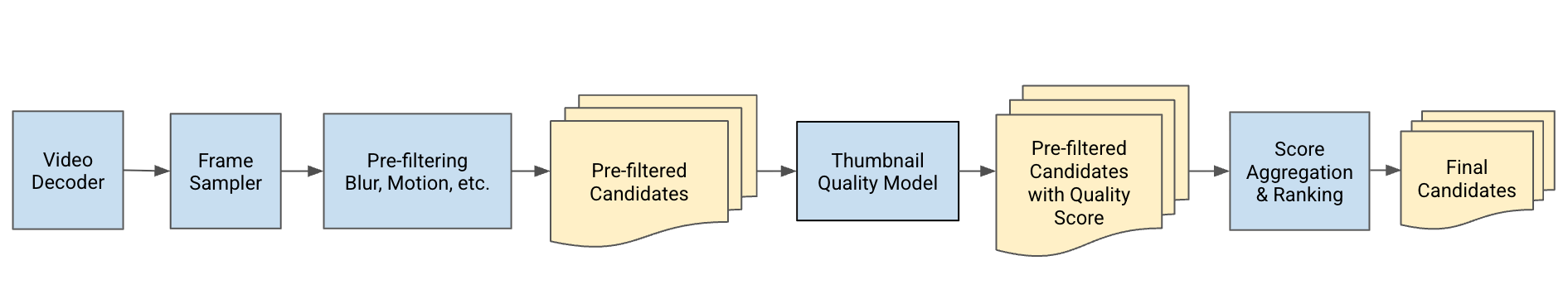}
  \caption{Thumbnail Candidate Selection Workflow}
  \Description{Multi-stage thumbnail candidate selection pipeline from video ingestion to final frame storage.}
\end{figure*}

\subsubsection{Temporal Segmentation and Pre-filtering} 
Candidate generation is embedded natively within the ingestion pipeline via a dedicated processing service. To guarantee semantic and visual diversity across the entire video timeline, the system partitions each upload into $K$ uniform temporal segments. Frames are sampled at a fixed frequency within each bounded segment and passed through a highly optimized, resource-efficient pre-filtering stage. This layer evaluates core visual heuristics, aggressively discarding frames exhibiting severe motion blur, compression artifacts and etc. Only candidate frames passing these strict visual thresholds are retained for advanced model scoring.

\subsubsection{Deep Quality Model Scoring} 
Following pre-filtering, each surviving candidate frame is evaluated by a deep thumbnail quality model to estimate its native visual appeal. The model architecture consists of a custom classification head integrated atop a pre-trained visual backbone, outputting dense semantic embeddings to subsequent projection layers optimized for binary classification. 

To build a robust training corpus without manual labeling, we programmatically derive positive examples from highly performing creator-curated thumbnails, applying strict filtering criteria based on historical engagement thresholds, trust and safety models, and low-level aesthetic metrics. Conversely, negative examples are sourced from random uncurated video frames exhibiting high semantic divergence from the creator asset within the embedding space, alongside frames classified as unsafe or visually degraded. Sampling across highly diverse content verticals ensures global platform representation.

To bridge the gap between deep semantic representation learning and the rigid computational limits of real-time ingestion, we employ a teacher-student knowledge distillation paradigm. While large-scale foundational Vision Transformers (ViT) \cite{Dosovitskiy2020} capture exceptional semantic nuances during offline experimentation, production resource profiling demonstrated that deploying a heavy ViT backbone at scale yields prohibitive inference latency and compute costs. Consequently, we employ a compact, highly optimized backbone for the online "student" model, while leveraging the frozen foundational ViT model as an offline "teacher." By training the student using a combined loss function that incorporates soft-label distillation from the teacher alongside hard ground-truth labels, the production model achieves near-parity predictive power while operating comfortably within target ingestion SLAs. 

The distillation topology is detailed in Figure 2.

\begin{figure*}[h]
  \centering
  \includegraphics[width=0.76\linewidth]{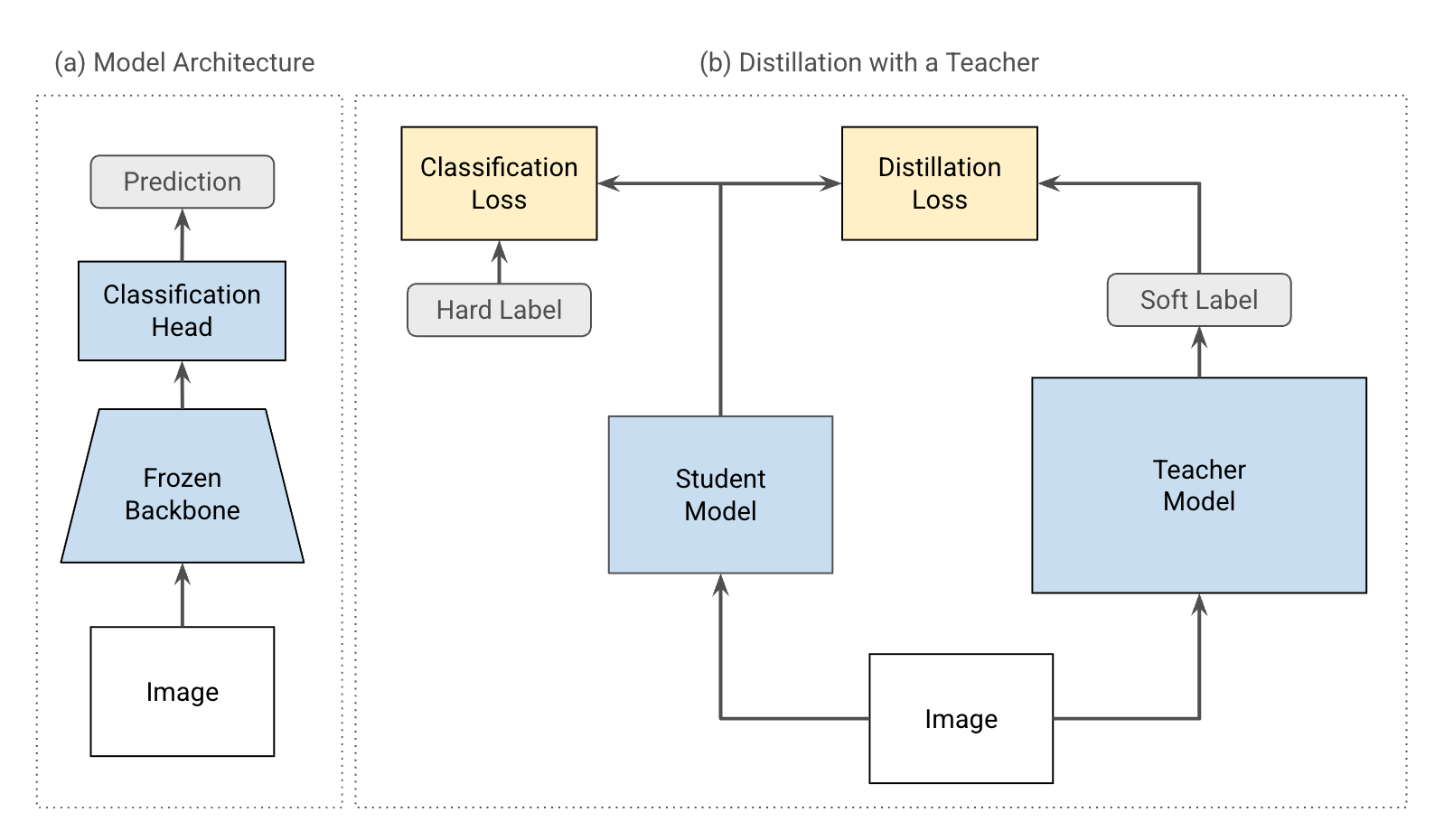}
  \caption{Teacher-student distillation topology for the thumbnail quality model. The offline teacher utilizes a heavy, frozen Vision Transformer backbone to generate soft targets, while the production student employs a highly streamlined backbone optimized for sub-millisecond serving.}
  \Description{Architecture diagram of teacher-student knowledge distillation for visual quality scoring.}
\end{figure*}

\subsubsection{Final Candidate Selection \& Prior Formulation} 
In the final extraction phase, the system computes a composite score for each candidate frame by fusing the deep quality model's scalar prediction with safety classifiers and platform-specific policy constraints. The pipeline deterministically extracts the single highest-scoring frame from each of the $K$ temporal segments. These $K$ distinct frames are materialized in backend storage alongside unique asset identifiers. Crucially, their normalized quality scores are directly mapped to instantiate the initial Beta hyper-parameters ($\alpha_{k,0}, \beta_{k,0}$), successfully bridging offline computer vision with online reinforcement learning.

\subsection{Real-Time Multi-Armed Bandit Selection}
The operational lifecycle of the real-time serving infrastructure is illustrated in Figure 3. The online MAB architecture dynamically selects the optimal frame at the backend layer and propagates the chosen asset directly to the client interface. User exposure events (impressions) and subsequent terminal actions (playbacks or skips) are continuously streamed back into the ingestion layer to close the reinforcement loop. By computing posterior distributions directly at serving time via Thompson Sampling, the platform continuously balances exploration variance with reward exploitation.

\begin{figure*}[h]
  \centering
  \includegraphics[width=0.76\linewidth]{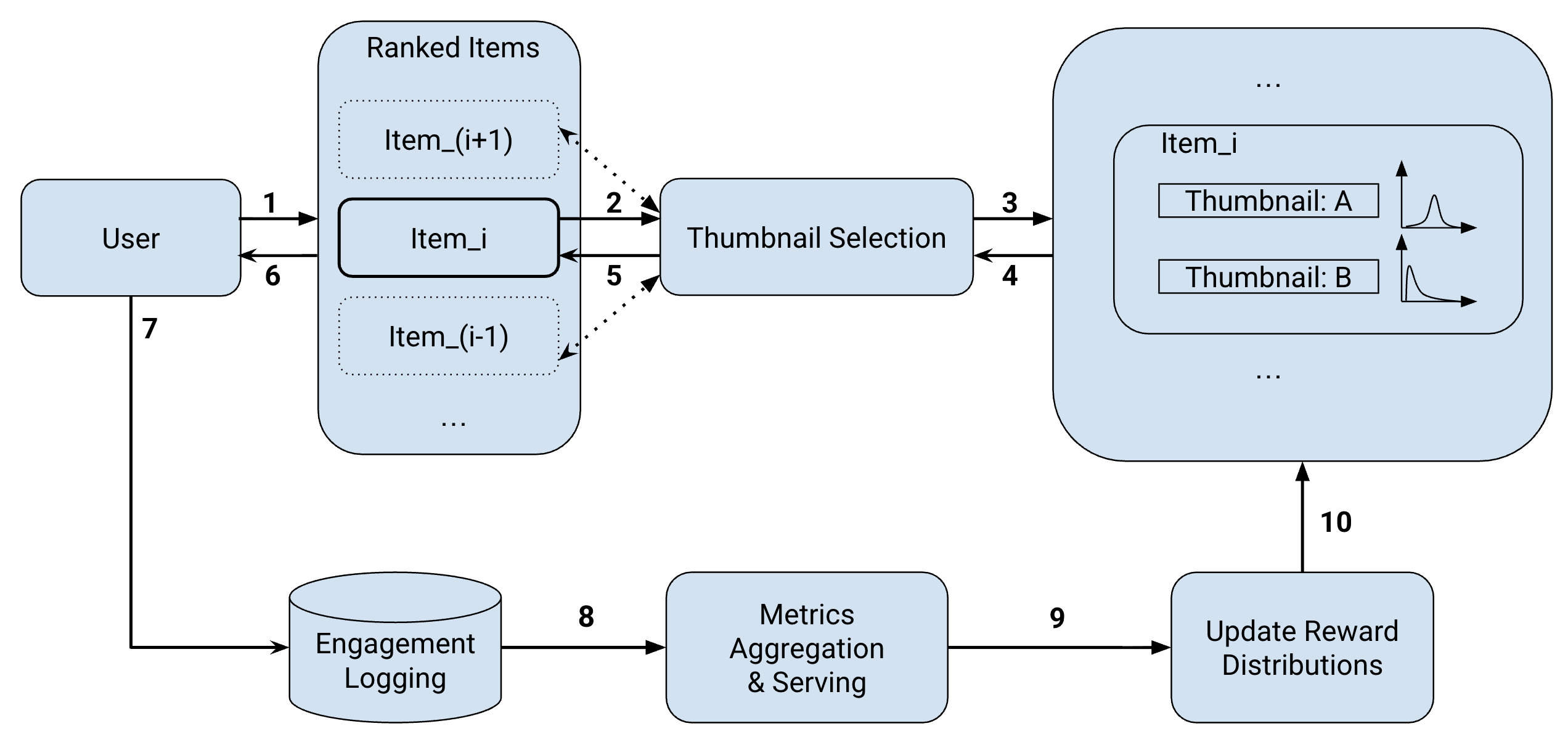}
  \caption{Real-Time Multi-Armed Bandit Selection Infrastructure}
  \Description{End-to-end systems architecture mapping real-time impression requests, Thompson Sampling evaluation, client rendering, and streaming feedback aggregation.}
\end{figure*}

\subsubsection{Distributed Thompson Sampling at Scale}
Rather than initializing exploration uniformly, candidate frames begin serving immediately weighted by their offline visual priors. As user interactions accumulate, the streaming feedback updates the underlying Beta reward distributions. At request time, backend nodes independently draw a random sample from the parameterized posterior distribution of each candidate arm. The system selects the frame yielding the maximum sampled value to render to the client. In the event of a sampling collision, a candidate is selected uniformly at random among the tied arms to preserve algorithmic fairness.

\subsubsection{Low-Latency Streaming Feedback Loop}
Operating an independent bandit model for every individual video across an $O(B)$ corpus requires extreme optimization of the feedback loop; minimizing round-trip attribution latency is vital to avoiding regret accumulation during early exploration. Impression and interaction telemetry streams directly into a distributed real-time metrics aggregation platform. 

A core system innovation of this architecture is the elimination of asynchronous, database-level materialization of posterior weights. Instead of continuously writing updated parameters to a persistent data store, the low-latency serving layer dynamically reconstructs the posterior Beta distributions in memory at request time using the aggregated statistics. This stateless serving paradigm drastically reduces backend write amplification, lowers storage overhead, and ensures the bandit updates reflect near real-time user feedback.

\subsubsection{End-to-End Infrastructure Integration}
Deploying this online learning paradigm platform-wide necessitated deep infrastructural modifications spanning the entire lifecycle of a request. The real-time bandit evaluation module is injected directly into the core recommendation stack immediately following the video ranking phase, such that the thumbnail reward collected is unbiased and not confounded by video ranking. By decoupling image optimization from static metadata generation, the infrastructure dynamically decorates recommended content items with their statistically optimal visual presentation in the critical path of the user request.

\section{Live Experiment}
Because a Multi-Armed Bandit (MAB) acts as an active decision-maker rather than a passive predictor, its efficacy must be evaluated online. Live experiments are essential for the algorithm to dynamically balance exploration and exploitation, enabling it to adapt to non-stationary user preferences in real time.

\subsection{Initial Experiment Setup}
In the production system, each video has a single-digit number of thumbnail candidates generated by the offline model at upload time. Online experiment was configured to test the dynamic thumbnail selection on a small percentage (e.g. 5\%) of all users. Safety guardrails were applied to operate on high-fidelity videos. The initial experiment applied uniform prior in Thompson Sampling in order to get a clean baseline for future iterations. Users were randomly assigned to the treatment or control group: treatment experience is the dynamic selection of thumbnail via multi-armed bandits; control experience is the default thumbnail from the middle section of the videos. After running the experiment for weeks, a two-sample $t$-test was performed between the two user groups in the latest 7-day period.

\subsection{Initial Experiment Result}
By optimizing the thumbnail displayed for short-form videos on the homepage, user value metrics increased statistically significantly. A strong learning effect was observed in the experiment stage, with an upward trend of success metrics over time. It happened naturally in bandit convergence, where the traffic mix shifted from exploration towards exploitation. Another step-function lift was observed after the feature was fully rolled out and measured in a holdback study. It suggested that the performance of multi-armed bandit algorithms is heavily traffic dependent - the higher coverage, the faster convergence. Aside from engagement lift, resource and latency costs are both neutral, demonstrating the algorithm is effective yet operationally inexpensive. Overall it grows Short-form user value metrics summarized in Table 1, including satisfied CTR (click-through rate), satisfied visits, satisfied users and satisfied engagement. These are online proprietary composite metrics reflecting prolonged user watch time and positive interactions (e.g., likes, saves) on the platform.
\begin{table*}
  \caption{Online Results for the Initial Launch of Thompson Sampling}
  \label{tab:freq}
  \begin{tabular}{lcccc}
    \toprule
    Metric & Satisfied CTR & Satisfied Visits & Satisfied Users & Satisfied Engagement\\
    \midrule
    Change & +1.72\% & +0.78\% & +0.68\% & +3.23\% \\
    CI (95\%) & [+1.67\%, +1.76\%] & [+0.75\%, +0.82\%] & [+0.65\%, +0.70\%] & [+3.10\%, +3.36\%] \\
    p-value & <0.0001 & <0.0001 & <0.0001 & <0.0001 \\
    \bottomrule
\end{tabular}
\end{table*}

\begin{figure}[h]
  \centering
  \includegraphics[width=0.75\linewidth]{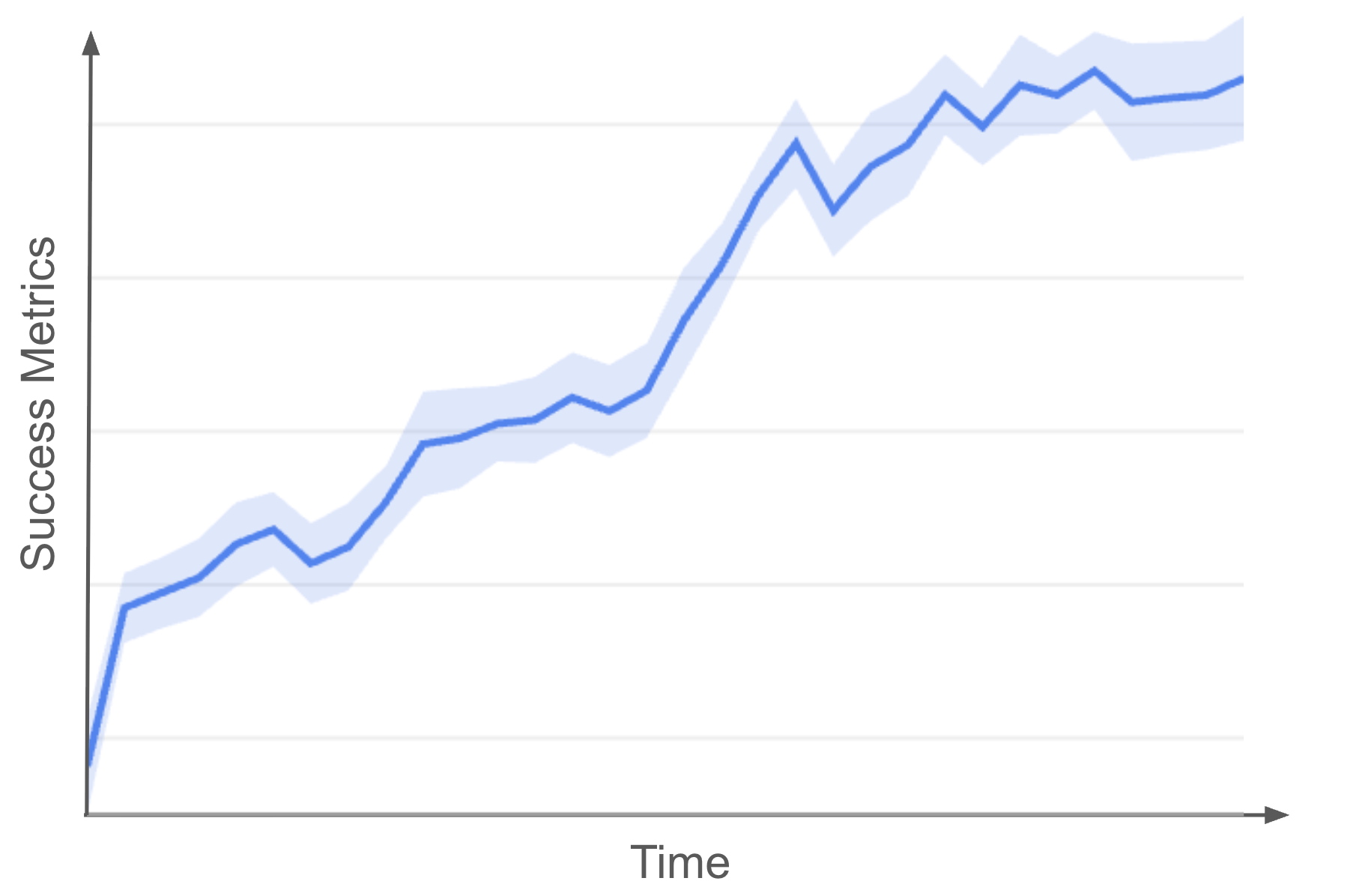}
  \caption{Metrics Trend Over Time}
  \Description{Metrics Trend Over Time}
\end{figure}

\subsection{Iterations}
Before launching the initial experiment of Thompson Sampling with uniform prior, iterations were made on the offline candidate selection model to select a better static thumbnail per video by mainly improving the deep quality model which yields modest positive gains across all core metrics compared to the static baseline. The same model was launched in production to generate a single-digit number of thumbnail candidates for downstream selection. As an alternative algorithm to Thompson Sampling, thumbnail randomization (randomly serve thumbnail candidates)  performed worse than the static baseline across the board. UCB was not tested due to the non-trivial latency of logging and aggregation in the data life cycle. As the optimal variant, Thompson Sampling with a uniform prior was fully deployed to maximize value delivery to users.

After the launch, iterations on the visual-quality priors and reward functions were performed to improve the algorithm. Because the performance of Thompson Sampling is traffic-dependent, a counterfactual variant with the same traffic size as the experiment variant was set up as the baseline for a fair comparison. Priors were initialized as the historical mean reward in each visual quality score bucket and confidence-scaled on an effective sample size, where the number of buckets and the effective sample size are both tunable parameters. Reward function was defined as a linear combination of reward actions in different time windows, where the weights of each time window are also tunable, aiming to adapt to recent user preferences faster. Despite the production challenge that does not allow exhaustive search of the parameters space, bundling the changes of visual quality-based prior initialization and recency-weighted reward function demonstrated incremental gains on the success metrics with statistical significance. The uplifts were mainly observed on the newly uploaded videos, which helped with the cold-start problem. As a sanity check, thumbnail randomization consistently performed worse than the baseline, further validating the efficacy of Thompson Sampling post-launch.

The iteration results are shown in Table 2.

\begin{table*}
  \caption{Online Results for the Iterations}
  \label{tab:freq}
  \begin{tabular}{lcccc}
    \toprule
    Iteration & Satisfied CTR & Satisfied Visits & Satisfied Users & Satisfied Engagement\\
    \midrule
    Static Thumbnail (baseline) & -- & -- & -- & -- \\
    Improved Candidate Selection & +0.08\% [-0.02\%, +0.18\%] & +0.04\% [-0.01\%, +0.10\%]& \textbf{+0.07\%} [+0.02\%, +0.12\%] & \textbf{+0.08\%} [+0.01\%, +0.16\%] \\
    Thumbnail Randomization  & \textbf{-0.86\%} [-0.94\%, -0.79\%] & \textbf{-0.33\%} [-0.39\%, -0.27\%] & \textbf{-0.28\%} [-0.35\%, -0.22\%] & \textbf{-0.97\%} [-1.18\%, -0.76\%]\\
    Thompson Sampling & \textbf{+1.72\%} [+1.67\%, +1.76\%] & \textbf{+0.78\%} [+0.75\%, +0.82\%] & \textbf{+0.68\%} [+0.65\%, +0.70\%] & \textbf{+3.23\%}  [+3.10\%, +3.36\%] \\
    \midrule
    Thompson Sampling (baseline) & -- & -- & -- & -- \\
    Thumbnail Randomization & \textbf{-1.79\%} [-1.85\%, -1.73\%] & \textbf{-0.86\%} [-0.92\%, -0.80\%] & \textbf{-0.74\%} [-0.78\%, -0.69\%] & \textbf{-2.97\%} [-3.14\%, -2.81\%] \\
    Prior and Reward Tuning & \textbf{+0.11\%} [+0.04\%, +0.18\%] & \textbf{+0.10\%} [+0.06\%, +0.13\%] & \textbf{+0.08\%} [+0.04\%, +0.12\%] & +0.16\% [-0.06\%, +0.39\%] \\
    \bottomrule
\end{tabular}
\end{table*}

\section{Conclusions and Future Work}
In this work, we presented an end-to-end framework for real-time thumbnail optimization deployed at a global $O(B)$ scale on a major short-form video platform. Our approach addresses two critical industry challenges: the pervasive absence of creator-curated artwork in short-form corpora, and the computational hurdles of operating independent Multi-Armed Bandits across billions of dynamic assets. By transitioning from static default frames to a data-driven framework, we successfully demonstrated that bridging offline computer vision with online reinforcement learning significantly enhances discovery and user engagement. Live platform-wide A/B experiments confirmed that this architecture delivers statistically significant gains across core user value metrics on the homepage.

The primary technical achievement of this system lies in making online explore-exploit computationally and experientially viable at massive scale. Rather than deploying standard cold-start bandits— which incur steep user-experience penalties by serving suboptimal frames during early exploration— our system initializes the exploration framework with informative priors derived from a specialized, automated, deep visual quality model. This allows the framework to bypass pure-exploration waste and focus impressions on highly viable candidates immediately. Furthermore, leveraging a Beta-Bernoulli conjugate setup for binary rewards enables lightweight, near real-time parameter updates. This ensures the system rapidly adapts to streaming user interactions without exceeding strict serving latency budgets. Overall, this framework provides a principled, fully automated, highly scalable blueprint for optimizing uncurated visual assets at an $O(B)$ scale.

Looking ahead, we identify several avenues to extend the capabilities and impact of this framework:

\begin{itemize}
\item \textbf{Contextual Personalization:} While our current deployment proves the strong efficacy of global-level optimization, viewer visual preferences are inherently heterogeneous. Transitioning to a contextual bandit \cite{li2010contextual, agrawal2013thompson, russo2018tutorial} architecture will allow the system to incorporate real-time user state and historical preferences, tailoring the visual gateway to individual user profiles.
\item \textbf{Cross-Surface Generalization:} We plan to expand this online optimization framework beyond the homepage to secondary discovery surfaces, such as dedicated search results and related-video feeds, ensuring cohesive visual presentation across the entire user journey.
\item \textbf{Closed-Loop Prior Refinement:} To better address the cold-start problem, we plan to explore a novel prior initialization approach that leverages Gemini prompt-inference for thumbnail quality assessment. To further accelerate bandit convergence, we aim to close the feedback loop by continuously injecting real-time bandit exploration telemetry back into the prompt-tuning process. By optimizing the visual quality priors against downstream bandit rewards, our system will dynamically refine its prompt configurations, ultimately yielding increasingly accurate Bayesian priors natively at ingestion time.
\end{itemize}

\bibliographystyle{ACM-Reference-Format}
\bibliography{main}

\end{document}